\documentclass[pdflatex,sn-nature,iicol]{sn-jnl} 

\usepackage{graphicx}%
\usepackage{multirow}%
\usepackage{amsmath,amssymb,amsfonts}%
\usepackage{amsthm}%
\usepackage{mathrsfs}%
\usepackage[title]{appendix}%
\usepackage{xcolor}%
\usepackage{textcomp}%
\usepackage{manyfoot}%
\usepackage{booktabs}%
\usepackage{algorithm}%
\usepackage{algorithmicx}%
\usepackage{algpseudocode}%
\usepackage{tabularx}
\usepackage{listings}%
\usepackage{float} 
\usepackage{tikz}
\usepackage{soul} 
\usepackage{color}
\usetikzlibrary{arrows.meta,positioning,fit,calc}

\theoremstyle{thmstyleone}%
\theoremstyle{thmstyletwo}%

\theoremstyle{thmstylethree}%

\begin{document}

\title[Article Title]{Beyond the Hype: Comparing Lightweight and Deep Learning Models for Air Quality Forecasting}


\author[1]{\fnm{Moazzam Umer} \sur{Gondal}}\email{moazzamumar22@gmail.com}

\author[1]{\fnm{Hamad} \sur{ul Qudous}}\email{hamad.ulqudous@nu.edu.pk}

\author[1]{\fnm{Asma Ahmad} \sur{Farhan}}\email{asma.ahmad@nu.edu.pk}

\affil[1]{\orgdiv{School of Computing}, \orgname{National University of Computer \& Emerging Sciences (FAST)}, \orgaddress{\city{Lahore}, \postcode{54000}, \country{Pakistan}}}


\abstract{
Accurate forecasting of urban air pollution is essential for protecting public health and guiding mitigation policies. While Deep Learning (DL) and hybrid pipelines dominate recent research, their complexity and limited interpretability hinder operational use. This study investigates whether lightweight additive models---Facebook Prophet (FBP) and NeuralProphet (NP)---can deliver competitive forecasts for particulate matter (PM$_{2.5}$, PM$_{10}$) in Beijing, China. Using multi-year pollutant and meteorological data, we applied systematic feature selection (correlation, mutual information, mRMR), leakage-safe scaling, and chronological data splits. Both models were trained with pollutant and precursor regressors, with NP additionally leveraging lagged dependencies. For context, two machine learning baselines (LSTM, LightGBM) and one traditional statistical model (SARIMAX) were also implemented. Performance was evaluated on a 7-day holdout using MAE, RMSE, and $R^2$. Results show that FBP consistently outperformed NP, SARIMAX, and the learning-based baselines, achieving test $R^2$ above 0.94 for both pollutants. These findings demonstrate that interpretable additive models remain competitive with both traditional and complex approaches, offering a practical balance of accuracy, transparency, and ease of deployment.
}



\maketitle

\section{Introduction}\label{sec1}

Air pollution remains a major global environmental and public health challenge. Rapid urbanization, industrial emissions, and motorized transport continue to degrade air quality in major cities. Among pollutants, fine particulate matter ($PM_{2.5}$) and coarse particles ($PM_{10}$) are of particular concern because they penetrate deeply into the respiratory system, causing respiratory and cardiovascular diseases \cite{bib1}, while also reducing visibility and contributing to climate change.

Cities such as Tehran and Beijing illustrate this global challenge, where frequent haze and high $PM_{2.5}$ levels highlight the difficulty of real-time management and the need for accurate forecasting \cite{bib2, bib3}.

Reliable forecasts are essential for public health and policy planning, as the WHO links over 4 million premature deaths in 2019 to air pollution \cite{bib1}. Accurate short-term predictions of $PM_{2.5}$ and $PM_{10}$ support timely warnings and interventions, yet forecasting in megacities remains difficult due to microclimates and complex pollutant interactions \cite{bib4}. Robust predictive frameworks also enable smart city functions such as adaptive traffic and health management \cite{bib4, bib5}.

Research spans classical statistical methods to Deep Learning (DL) and hybrid decompositions. LSTM architectures have shown strong short-term performance, particularly when combined with meteorological drivers or hyperparameter tuning \cite{bib6}. Decomposition–reconstruction pipelines (e.g., SSA/VMD plus LSTM or gradient boosting) improve fidelity by isolating multi-scale components, outperforming monolithic deep models \cite{bib5, bib7}. More recent hybrid deep learning frameworks, such as BiGRU–1DCNN models for spatio-temporal $PM_{2.5}$ forecasting, have further advanced prediction accuracy in urban settings \cite{bibr1-1}. Reviews emphasize the complementary value of such hybrids and the growing role of cloud–edge ecosystems \cite{bib4, bib7}. In parallel, Federated Learning (FL) preserves data locality across agencies while training global or personalized predictors. Surveys document the promise and challenges of FL, with applications including convolutional–recurrent models \cite{bib9}, UAV-assisted prediction \cite{bib10}, and graph-based FL for non-Euclidean spatial structures \cite{bib11}. While these approaches achieve high accuracy, they often introduce operational complexity and reduce transparency for policymakers.

Forecasting $PM_{2.5}$/$PM_{10}$ in Beijing exemplifies several challenges: nonstationary signals with multi-scale seasonality, pollutant–pollutant coupling, and meteorological regimes that shift across months. Data gaps from sensor downtime and episodic haze require robust preprocessing and careful experimental design to prevent information leakage. Feature selection also remains crucial, as recent work highlights mutual-information filtering for spatio-temporal networks \cite{bib12}.

In this study, we adopt a transparent and operational modeling choice: primarily comparing two additive time-series frameworks with exogenous regressors—Facebook Prophet and NeuralProphet—for hourly $PM_{2.5}$ and $PM_{10}$ forecasting in Beijing. Both encode trend–seasonality structure and accept external drivers; NeuralProphet additionally incorporates lagged dependencies to capture short-term autocorrelation while remaining interpretable. To contextualize their performance, we also benchmark against two widely used baselines: Long Short-Term Memory (LSTM) networks and LightGBM, representing DL and gradient boosting paradigms. We curate meteorological and pollutant covariates via mutual information and mRMR, enforce leakage-safe scaling (fit only on training regressors), and evaluate using a held-out validation set and a last-week test horizon mimicking deployment.

Our contributions are threefold:

\begin{itemize}
    \item A leakage-safe methodology combining additive forecasting with modern feature selection, supporting multivariate air-quality prediction while offering a compact, interpretable alternative to DL.
    \item A controlled comparison of FB Prophet and NeuralProphet with exogenous drivers, benchmarked against LSTM, SARIMAX and LightGBM, and evaluated across training, validation, and test windows using MAE, RMSE, and R² under a realistic last-week forecast protocol.
    \item Actionable insights showing that carefully engineered additive models can deliver competitive accuracy while maintaining simplicity, interpretability, and auditability—qualities essential for city agencies and health stakeholders.
\end{itemize}

\section{Related Work}

Research on air-quality forecasting has progressed from classical statistical baselines to Machine Learning (ML), Deep Learning (DL), and hybrid frameworks. Early models emphasized simplicity and interpretability but struggled with nonlinear, multivariate behavior. As sensing networks expanded and computation became cheaper, ML and DL methods advanced the state of the art by learning richer temporal and spatial dependencies. Hybrid pipelines further combined decomposition, feature selection, and model fusion to boost robustness and accuracy. Across this evolution, a key tension persists between predictive performance and the transparency required for policy use \cite{bib13,bib16,bib4}.

Classical statistical models such as ARIMA/SARIMA and multiple linear regression have long been applied to PM$_{2.5}$, PM$_{10}$, and AQI forecasting because they are data-efficient and interpretable \cite{bib13}. Facebook Prophet revitalized additive regression as a modern baseline; studies reported that Prophet captured recurring seasonal patterns and often outperformed tuned SARIMA for short horizons \cite{bib14,bib15}. However, fixed parametric structures limit adaptability to regime shifts and high-dimensional inputs, as urban pollution is influenced by nonlinear meteorology and variable emissions \cite{bib16}. While statistical approaches remain valuable for transparency and medium-range seasonality, their rigidity motivated the move toward ML and DL \cite{bib13,bib14,bib15,bib16}.

ML methods offer flexibility for nonlinearities, heterogeneous covariates, and spatial transfer. Tree ensembles and margin-based learners are widely used: LightGBM leveraged high-dimensional spatio-temporal features for next-day PM$_{2.5}$ across 35 Beijing stations \cite{bib17}, while evaluations across 23 Indian cities highlighted ensemble strength but also year-to-year drift \cite{bib18}. End-to-end ML pipelines emphasize preprocessing, feature importance, and careful train/test design for Random Forest, SVM, and boosting methods \cite{bib19}. Spatial generalization has been demonstrated via statewide PM$_{2.5}$ surfaces with auxiliary covariates \cite{bib20}. ML has also been adapted to atypical regimes such as mass rallies using temporally weighted multitask learning \cite{bib21}. Recent work further explores robust ML rules for pollution trend detection, integrating PCA, clustering, and correspondence analysis to identify multimodal pollution regimes \cite{bibr2-1}. Comparative studies in Gulf-region meteorology showed that ML remains competitive with both DL and Prophet for one- to seven-day horizons \cite{bib15}. Complementary findings show that optimized ML regressors—particularly SVR, LightGBM, and ensemble stacks—can yield highly accurate forecasts when combined with Bayesian optimization and systematic preprocessing \cite{bibr2-2}. Surveys echo these findings: while DL often leads on accuracy, ML baselines retain reliability and integrate naturally with IoT-centric sensing for calibration and forecasting \cite{bib16,bib4,bib19,bib20}.

DL now dominates for capturing long-range dependencies and complex multivariate dynamics. RNN variants—especially LSTM and Bi-LSTM—have been applied at continental scale using attention mechanisms and heterogeneous data sources, improving event sensitivity and accuracy \cite{bib22}. Hyperparameter tuning strongly affects performance, with optimized Bi-LSTM models outperforming untuned baselines \cite{bib23}. Continuous-time formulations via Neural ODEs address irregular sampling and nonstationarity, yielding gains over conventional LSTM \cite{bib24}. DL also accommodates atypical drivers: traffic and noise data improved PM$_{10}$ forecasts in Skopje, with noise contributing more than traffic \cite{bib25}. Hybrid DL pipelines combining decomposition and learning (e.g., Bi-LSTM with LightGBM) achieved further gains \cite{bib26}. Despite progress, DL requires larger datasets and compute budgets and remains less transparent for operational agencies \cite{bib22,bib23,bib24,bib25,bib26}.

Hybrid and ensemble strategies integrate complementary strengths across decomposition, architecture fusion, and feature filtering. Decomposition-based hybrids (e.g., SSA or STL followed by specialized learners) improve stability by modeling trend/seasonal/noise components separately before recombination, achieving lower RMSE/MAE/SMAPE than vanilla deep baselines \cite{bib26,bib27}. Recent studies have continued to enhance hybrid architectures by integrating recurrent, convolutional, and boosting components. For example, BiGRU–1DCNN frameworks have shown strong spatio-temporal learning performance across urban $PM_{2.5}$ networks \cite{bibr1-1}, while hybrid RNN–BiGRU models with advanced imputation methods improved robustness under missing data \cite{bibr1-2}. Similarly, WaveNet-inspired CNN–BiLSTM–XGBoost models achieved notable gains in multi-city forecasting accuracy \cite{bibr1-3}. Architectural fusion further combines global attention with local memory (Transformer–LSTM) and uses metaheuristics for hyperparameter tuning, delivering consistent gains across Chinese cities \cite{bib28}. At the systems level, federated learning (FL) enables privacy-preserving collaboration under non-IID data and client drift, while IoT–AI stacks connect low-cost sensors to ML/DL modules for calibration and forecasting \cite{bib30,bib4}. Collectively, these hybrids enhance accuracy and robustness but also increase implementation complexity and operational overhead \cite{bib26,bib27,bibr1-1,bibr1-2,bibr1-3,bib28,bib29,bib30,bib4}.

Despite this progress, two gaps persist. First, much work optimizes for accuracy through increasingly complex pipelines, with less emphasis on interpretability and ease of deployment—qualities crucial for policy response \cite{bib4,bib16}. Second, systematic assessments of modern additive time-series models for air-pollution forecasting remain limited. In particular, Facebook Prophet and NeuralProphet encode trend and seasonality explicitly, accept external regressors, and (for NeuralProphet) permit lagged dependencies, offering lightweight yet interpretable alternatives to heavy DL pipelines \cite{bib14,bib15}. This study addresses these gaps through a head-to-head evaluation of Prophet and NeuralProphet for PM$_{2.5}$/PM$_{10}$ forecasting, benchmarking accuracy while emphasizing transparency and practical usability.

The remainder of this paper is organized as follows. Section~\ref{sec:methodology} describes the dataset, preprocessing steps, and feature selection. Section~\ref{sec:implementation} outlines model implementation details. Section~\ref{sec:results} presents results and comparative analyses, while Section~\ref{sec:discussion} discusses key implications. Section~\ref{sec:conclusion} concludes the study and outlines future work.

\begin{table*}[!t]
\caption{Condensed summary of representative literature (refs \cite{bib4,bib13,bib14,bib15,bib16,bib17,bib18,bib19,bib20,bib21,bib22,bib23,bib24,bib25,bib26,bib27,bibr1-1,bibr1-2,bibr1-3,bib28,bib29,bib30}).}
\label{tab:lr}
\setlength{\tabcolsep}{4pt}
\renewcommand{\arraystretch}{1.05}
\footnotesize
\begin{tabularx}{\textwidth}{|>{\centering\arraybackslash}p{0.6cm}|p{2.4cm}|p{2.3cm}|X|}
\hline
{\scriptsize \#} & \textbf{Task/Target} & \textbf{Data/Region} & \textbf{Method \& key highlights} \\
\hline

\cite{bib4} & IoT AQM + AI (review) & PRISMA 2016–2024 & Systematic review; taxonomy (\emph{imputation, calibration, anomalies, AQI, short-term forecasting}); stresses data quality, scalability and deployment gaps. \\ \hline

\cite{bibr1-1} & PM$_{2.5}$ forecasting (spatio-temporal hybrid) & Delhi, India (2018–2023) & \textbf{BiGRU–1DCNN hybrid} integrates recurrent and convolutional layers for spatio-temporal learning across 28 CPCB stations; achieves lowest RMSE/MAE and strong seasonal trend capture. \\ \hline

\cite{bib13} & AQI forecast; stat.~vs NN & Beijing, China & \textbf{ARIMA} + \textbf{ANN}; linear vs nonlinear contrast; hybridization aids robustness; seasonality handling highlighted. \\ \hline

\cite{bib16} & Review of AI techniques for air-pollution forecasting & Global survey (2000–2020) & Comprehensive review of emerging \textbf{AI/ML methods} (SVM, ensembles, DL). Highlights evolution from statistical to hybrid models. Serves as a knowledge base for selecting forecasting strategies. \\ \hline

\cite{bib14} & Uni-variate short-horizon forecast (pollutants/AQI) & City station series & \textbf{SARIMA} vs \textbf{Prophet}; additive seasonality/log-transform improve short-horizon accuracy; interpretable baseline for ML/DL. \\ \hline

\cite{bib15} & PM$_{2.5}$/PM$_{10}$ comparison & UAE monitoring sites & \textbf{DT/RF/SVR} vs \textbf{CNN/LSTM} vs \textbf{Prophet}; ML competitive across 1–7 day horizons; notes climate-specific transfer. \\ \hline

\cite{bib17} & Next-day PM$_{2.5}$/AQI (station-level) & 35 Beijing stations & \textbf{LightGBM} with rich spatio-temporal covariates; systematic feature exploration; strong non-DL baseline, interpretable importances. \\ \hline

\cite{bib18} & AQI classification/regression & 23 Indian cities, 6 yrs (CPCB) & Multi-model \textbf{ML} benchmark (NB/SVM/XGBoost); correlation-based feature selection; evidences data drift \& generalization issues. \\ \hline

\cite{bib19} & AQI events at 1/8/24h & Event-focused datasets & End-to-end \textbf{ML} pipeline (RF/AdaBoost/SVM/ANN/stacking); preprocessing \& horizon design matter; ensembles often strongest. \\ \hline

\cite{bib20} & Spatially continuous PM$_{2.5}$ & New York State, USA & Statewide \textbf{predictive mapping} from monitors + auxiliary covariates; improved spatial generalization for exposure assessment. \\ \hline

\cite{bib21} & Multi-pollutant during mass rallies & Event-driven series & \textbf{Temporally weighted multitask} learner (SVR/TSVR-style); reweights in time, shares across pollutants; handles distribution shift. \\ \hline

\cite{bibr2-1} & Pollution trend detection (robust ML) & Southern China, Hong Kong, Macau (multi-year) & EstiMax and SMA algorithms integrate PCA, clustering, EM and correspondence analysis to identify multimodal pollution regimes and seasonal patterns across large spatio-temporal datasets. \\ \hline

\cite{bibr2-2} & ML forecasting (optimization + ensembles) & Hourly sensor data (1 year) & Comparative study of ten ML regressors; Bayesian optimization and stacking markedly improve SVR, LightGBM and boosting models, achieving high-precision short-term forecasts. \\ \hline

\cite{bib22} & Daily PM$_{2.5}$ estimation & CONUS (USA) & \textbf{Bi-LSTM + Attention}; fuses in-situ, satellite, wildfire smoke; better extreme-event capture; high-res national surfaces. \\ \hline

\cite{bib23} & PM$_{2.5}$ estimation & Multi-feature stations & \textbf{Bi-LSTM} with \textbf{Osprey} HPO; tuned deep models outperform untuned baselines; sensitivity to architecture/HPO. \\ \hline

\cite{bib24} & Short-horizon PM$_{2.5}$ (1–8 h) & Irregular sampling & \textbf{Neural-ODE} variants; continuous-time dynamics; \emph{2.9–14.1\%} gains vs LSTM; robust under irregular intervals. \\ \hline

\cite{bib25} & PM$_{10}$ with exogenous signals & Skopje (urban) & LSTM-family with \textbf{traffic} \& \textbf{noise}; noise aids prediction; traffic not always dominant—feature-design caution. \\ \hline

\cite{bib26} & Hourly AQI, 1-step/multi-step & Beijing, China & \textbf{SSA} $\rightarrow$ \textbf{Bi-LSTM} (per component) $\rightarrow$ \textbf{LightGBM} stack; decomposition stabilizes; stacking boosts generalization. \\ \hline

\cite{bib27} & PM$_{2.5}$ via component tailoring & Five Chinese cities & \textbf{STL} + component-specific learners; “right tool per component” improves MSE/MAE/MAPE/R$^2$; robust to nonstationarity. \\ \hline

\cite{bibr1-2} & PM$_{2.5}$ forecasting with missing-data handling & Lucknow, India (2020–2023) & \textbf{nRI RNN–BiGRU hybrid} combines novel random imputation with bidirectional recurrent modeling; improves robustness and accuracy under sensor data gaps. \\ \hline

\cite{bibr1-3} & Global urban PM$_{2.5}$ forecasting (hybrid deep ensemble) & Multi-city (US embassies, 2017–2024) & \textbf{1DCNN–BiLSTM–XGBoost (WaveNet-based)} hybrid achieves state-of-the-art accuracy via residual correction; effective for diverse climatic and pollution conditions. \\ \hline

\cite{bib28} & PM$_{2.5}$ model fusion & Central/Western China & \textbf{Transformer} $\oplus$ \textbf{LSTM} with \textbf{PSO} tuning; combines global attention + local memory; consistent gains. \\ \hline

\cite{bib30} & AQI under privacy constraints & FL/edge settings & \textbf{Multi-Model FL} survey; non-IID, client drift, comms costs; pipeline patterns \& open challenges. \\ \hline

\cite{bib29} & PM$_{2.5}$ with input curation & Multi-scenario eval. & \textbf{MI + AID} feature filtering $\rightarrow$ \textbf{Bayes-optimized STCN}; reduces redundancy; stabilizes training. \\ \hline

\end{tabularx}
\vspace{-3mm}
\end{table*}

\section{Methodology}\label{sec:methodology}

\begin{figure}[htbp]
    \centering
    \includegraphics[width=\linewidth]{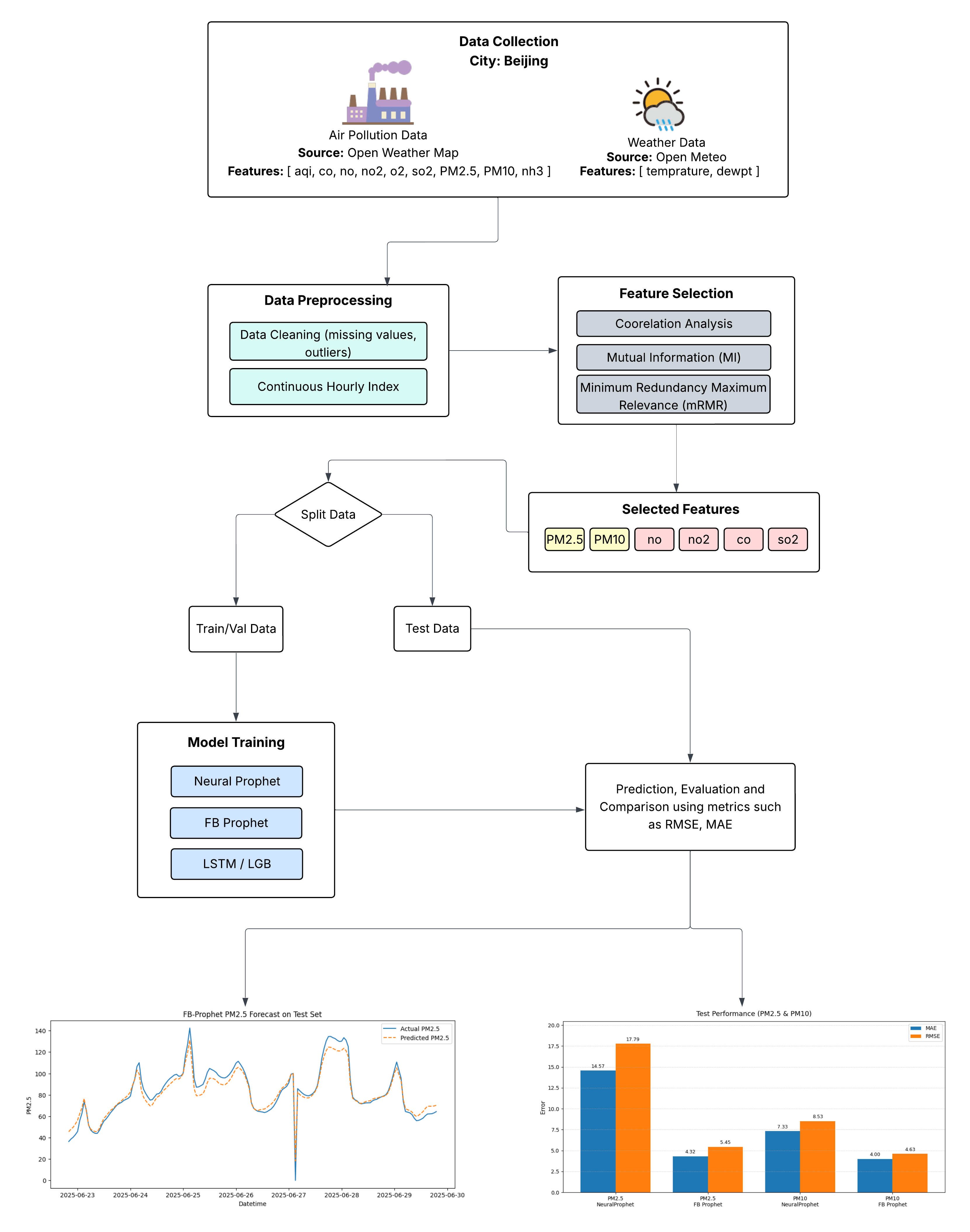}
    \caption{The figure illustrates the sequential workflow adopted in this research, covering data collection, preprocessing, feature selection, data partitioning, model training, and performance evaluation for air quality forecasting.}
    \label{fig:framework}
\end{figure}

The study followed a structured workflow comprising exploratory analysis, preprocessing, feature selection, chronological splitting, model training, and evaluation. For comparison, five data-driven approaches—Facebook Prophet (FBP), NeuralProphet (NP), Long Short-Term Memory (LSTM), SARIMAX (Seasonal ARIMA with Exogenous Factors) and LightGBM—were benchmarked alongside the traditional SARIMAX model. All models used the same preprocessed inputs for fairness. The workflow is shown in Fig.~\ref{fig:framework}.

\subsection{Data and Preprocessing}

An hourly time series was compiled for Beijing covering December~2020 to June~2025. Pollutant concentrations (PM$_{2.5}$, PM$_{10}$, CO, NO, NO$_2$, SO$_2$, O$_3$, NH$_3$) were obtained from the \textbf{OpenWeather} API \cite{bibOpenWeather}, while meteorological variables (temperature and dew point) were sourced from \textbf{Open-Meteo} \cite{bibOpenMeteo}. Both services provide model-based gridded estimates at user-specified coordinates rather than measurements from identifiable monitoring stations or sensors; OpenWeather blends satellite observations, surface stations, radar inputs, and global numerical weather prediction models, while Open-Meteo aggregates multi-model forecasts and reanalysis products from national weather agencies such as ECMWF and NOAA. Consequently, the exact number or spatial distribution of stations, sensor type and accuracy specifications are not published by either provider.

A continuous hourly index was enforced, but no missing timestamps or missing values were present after alignment, so no imputation was required. Anomalies were removed using z-scores, and all variables were retained at hourly resolution before feature selection.

\subsection{Feature Selection}

To identify informative yet nonredundant regressors, correlation, mutual information (MI), and minimum redundancy–maximum relevance (mRMR) were applied.  

The Pearson correlation between variables $X$ and $Y$ is  
\begin{equation}
\rho_{X,Y} = \frac{\mathrm{Cov}(X,Y)}{\sigma_X \sigma_Y},
\end{equation}
where $\mathrm{Cov}(X,Y)$ is the covariance between $X$ and $Y$, and $\sigma_X$, $\sigma_Y$ their standard deviations, capturing linear dependencies.

Mutual information quantifies nonlinear associations as  
\begin{equation}
I(X;Y) = \sum_{x}\sum_{y} p(x,y)\,\log \frac{p(x,y)}{p(x)p(y)},
\end{equation}
where $p(x,y)$ is the joint probability of $X$ and $Y$, and $p(x)$, $p(y)$ are their marginals.

Finally, mRMR balances informativeness and redundancy:  
\begin{equation}
\text{mRMR} = \max_{f \in F \setminus S} \Big[ I(f;y) - \tfrac{1}{|S|}\sum_{s \in S} I(f;s) \Big],
\end{equation}
where $F$ is the full feature set, $S$ the subset of selected features, and $I(f;y)$, $I(f;s)$ denote mutual information with the target and another feature, respectively.

This ensured the final feature set remained both relevant to PM$_{2.5}$/PM$_{10}$ and diverse across predictors.

\subsection{Data Splitting and Scaling}

Data were split chronologically into 80\% training, 10\% validation, and the final 7 days (168 hours) for testing, emulating real deployment. Regressors were standardized as $x'=(x-\mu_{\text{train}})/\sigma_{\text{train}}$ using training statistics to prevent leakage, while targets remained unscaled for interpretability.

\subsection{Models}

Five forecasting models were compared, representing additive, deep, and ensemble paradigms.

\paragraph{Facebook Prophet (FBP):}  
Prophet decomposes the series into trend, seasonality, and external regressors:
\begin{equation}
y(t) = g(t) + s(t) + h(t) + \epsilon_t,
\end{equation}
where $g(t)$ is trend which captures piecewise linear trends with changepoints, $s(t)$ is seasonality which models multi-period seasonality using Fourier terms, $h(t)$ are external regressors which incorporates pollutant and precursor covariates as additive effects, and $\epsilon_t$ noise. Its structural design enables interpretable long-term and recurring patterns.

\paragraph{NeuralProphet (NP):}
NP extends Prophet with autoregressive and lagged components:
\begin{equation}
y(t) = T(t) + S(t) + E(t) + A(t) + F(t) + L(t),
\end{equation}
where $T(t)$, $S(t)$, and $E(t)$ represent trend, seasonality, and event effects; $A(t)$, $F(t)$, and $L(t)$ capture autoregressive, future, and lagged regressors. This hybrid design combines Prophet-style decomposition with short-term temporal learning.

\paragraph{Long Short-Term Memory (LSTM):}  
LSTM models sequential dependencies, with cell state $c_t$ evolving as
\begin{equation}
c_t = f_t \odot c_{t-1} + i_t \odot \tilde{c}_t,
\end{equation}
where $f_t$ and $i_t$ are forget and input gates, and $\tilde{c}_t$ is the candidate update, enabling long-term pattern retention. This gating structure enables the network to retain multi-step temporal memory and handle nonlinear pollutant dynamics.

\paragraph{SARIMAX:}
SARIMAX extends ARIMA by incorporating seasonality and exogenous regressors:
\begin{equation}
y_t = c + \phi_1 y_{t-1} + \theta_1 \varepsilon_{t-1} + \Phi_1 y_{t-s} + \Theta_1 \varepsilon_{t-s} + \beta X_t + \varepsilon_t,
\end{equation}
where $y_t$ is the pollutant concentration at time $t$, $\varepsilon_t$ white noise, $X_t$ exogenous regressors, $\phi_1$, $\Phi_1$ autoregressive terms, $\theta_1$, $\Theta_1$ moving-average terms, and $s$ the seasonal period.
$(\phi_1,\theta_1)$ and $(\Phi_1,\Theta_1)$ capture non-seasonal and seasonal autoregressive and moving-average dynamics, and $X_t$ introduces external pollutant predictors. This provides a transparent baseline for linear and seasonal dependencies.

\paragraph{LightGBM:}  
LightGBM, a gradient boosting framework, updates prediction as
\begin{equation}
\hat{y}_i^{(m)} = \hat{y}_i^{(m-1)} + \eta \, T_m(x_i),
\end{equation}
where $T_m$ is the new decision tree and $\eta$ the learning rate. It builds trees using leaf-wise splitting with histogram-based optimization, allowing efficient modeling of nonlinear interactions among pollutant and meteorological features.

\subsection{Evaluation}

Performance was measured using Mean Absolute Error (MAE), Root Mean Squared Error (RMSE), and coefficient of determination ($R^2$):  
\begin{align}
\text{MAE} &= \tfrac{1}{N}\sum_{i=1}^{N} |y_i-\hat{y}_i|, \\
\text{RMSE} &= \sqrt{\tfrac{1}{N}\sum_{i=1}^{N} (y_i-\hat{y}_i)^2}, \\
R^2 &= 1 - \frac{\sum_{i}(y_i-\hat{y}_i)^2}{\sum_{i}(y_i-\bar{y})^2}.
\end{align}
Here, $y_i$ and $\hat{y}_i$ are observed and predicted pollutant concentrations, $\bar{y}$ the mean of observed values, and $N$ the number of samples. MAE and RMSE measure absolute and squared deviations, while $R^2$ reflects explained variance. These metrics are widely adopted in air-quality forecasting because they provide interpretable error magnitudes and allow direct comparison with prior studies. Additional metrics such as MAPE or SMAPE were avoided to prevent distortion under near-zero pollution values common in clean-hour periods

Metrics were computed for all subsets, with the final 7-day test set simulating deployment. Visual comparison of predicted versus actual values complemented the quantitative results.

\section{Implementation}\label{sec:implementation}

The workflow was implemented in Python using open-source libraries. Data handling and preprocessing employed \texttt{pandas} and \texttt{scikit-learn}, feature selection used \texttt{pymrmr}, and forecasting relied on \texttt{prophet} and \texttt{neuralprophet}. NeuralProphet was trained on GPU via \texttt{PyTorch Lightning}, while LSTM and LightGBM models were developed in \texttt{TensorFlow/Keras} and \texttt{lightgbm}. Visualizations were produced with \texttt{matplotlib}, and random seeds ensured reproducibility.

\subsection{Dataset and Preparation}

Hourly pollutant and meteorological time series for December~2020 to June~2025 were retrieved from the same sources described in Section ~\ref{sec:methodology}. Targets were PM$_{2.5}$ and PM$_{10}$, with regressors including CO, NO, NO$_2$, SO$_2$, and NH$_3$. After merging the datasets, a continuous hourly index was confirmed, anomalies were filtered using z-scores, and no missing values were detected. StandardScaler was applied to regressors before model training.

\begin{figure}[htbp]
    \centering
    \includegraphics[width=\linewidth]{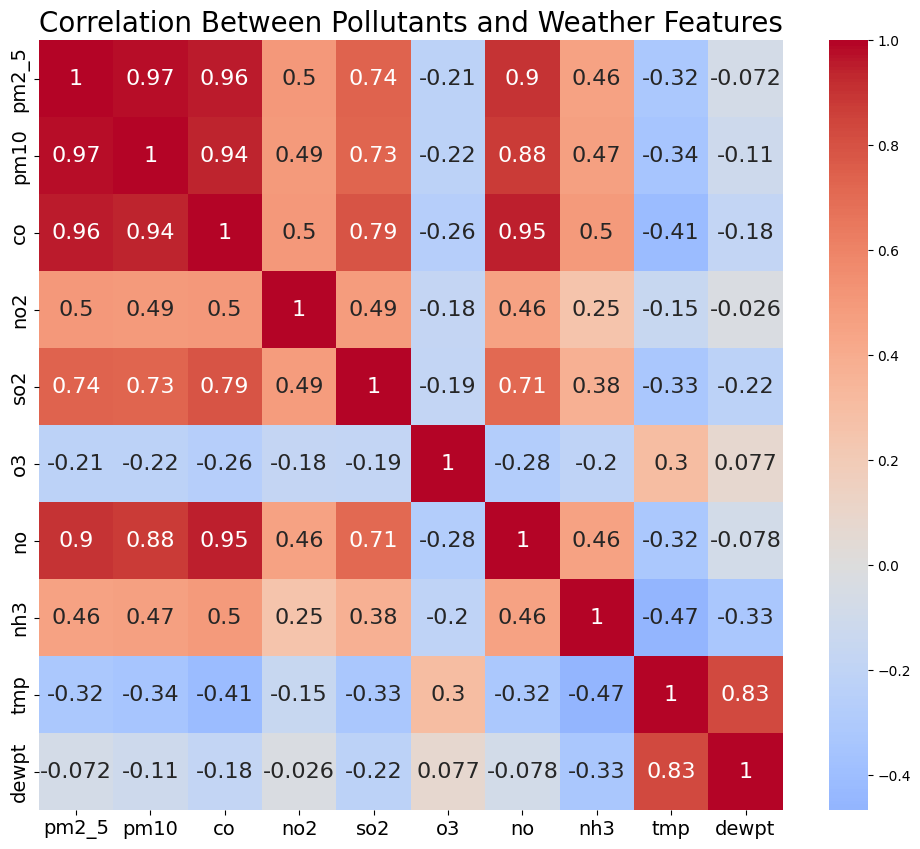}
    \caption{The heatmap visualizes pairwise relationships among variables, highlighting strong associations between particulate matter and gaseous precursors.}
    \label{fig:corr_heatmap}
\end{figure}

\begin{figure}[htbp]
    \centering
    \includegraphics[width=\linewidth]{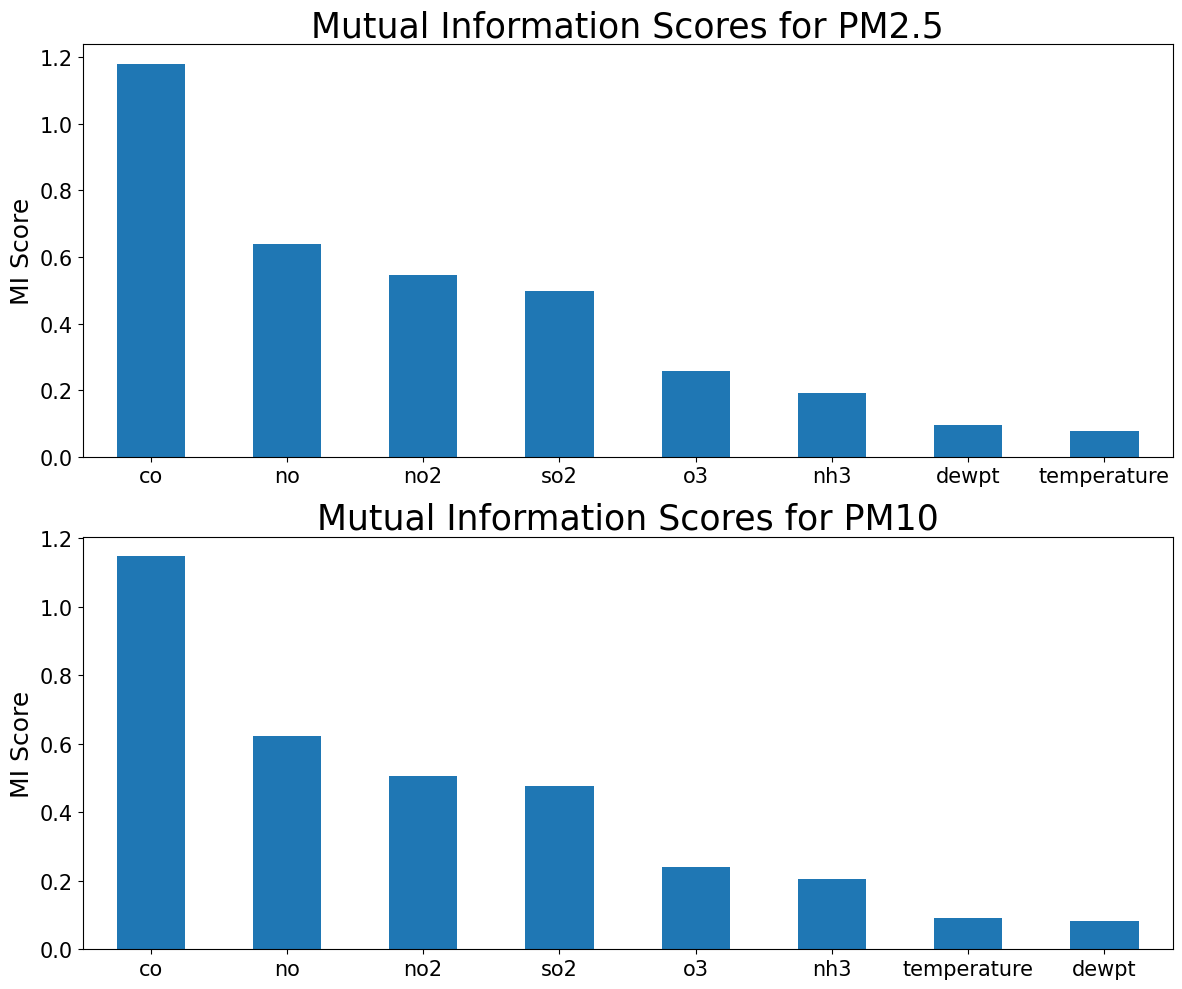}
    \caption{This plot ranks features by their nonlinear relevance to the target pollutants, guiding the selection of informative predictors.}
    \label{fig:mi_bar}
\end{figure}

Correlation analysis (Fig.~\ref{fig:corr_heatmap}) revealed strong relationships between particulates and gaseous precursors, while Mutual Information confirmed CO, NO, NO$_2$, and SO$_2$ as key predictors (Fig.~\ref{fig:mi_bar}). mRMR further reduced redundancy and showed cross-dependence between PM$_{2.5}$ and PM$_{10}$; hence, each was used as a regressor for the other. Gaseous pollutants were retained as indicators of combustion activity, whereas meteorological variables had weaker predictive influence and were not prioritized.  

The final feature--target sets were:
- PM$_{2.5}$ forecast with \{PM$_{10}$, NO, NO$_2$, CO, SO$_2$\},  
- PM$_{10}$ forecast with \{PM$_{2.5}$, NO, NO$_2$, CO, SO$_2$\}.  

\subsection{Model Configurations}

\textbf{Facebook Prophet (FBP):} Weekly and yearly seasonalities were enabled, consistent with established practice for urban air-quality series, while daily cycles were disabled due to weak diurnal structure in exploratory analysis. These settings follow recommended defaults in the Prophet literature and provided stable validation performance. Pollutant and precursor variables were included as external regressors, and forecasts were generated for the 168-hour horizon.

\textbf{NeuralProphet (NP):} Weekly and yearly seasonalities were retained, and \texttt{n\_lags}=7 was selected to capture short-term autocorrelation without substantially increasing model complexity. This choice aligns with both documentation guidance and preliminary validation experiments showing no improvement with larger lag windows. Training used 50 epochs with validation monitoring.

\textbf{SARIMAX:} The configuration $(1,0,1)(1,0,1,24)$ reflects a standard hourly-seasonal form often adopted for pollutant time series and was validated through ACF/PACF inspection. This provided a stable baseline while avoiding excessive parameterization.

\textbf{LSTM:} A single hidden layer with 64 units was used to balance model capacity and overfitting risk. Deeper or wider architectures were tested but did not yield consistent validation improvements, so a compact configuration was retained for stability. The model used ReLU activation, the Adam optimizer, and early stopping over 50 epochs.

\textbf{LightGBM:} A limit of 500 boosting rounds was chosen as a commonly adopted and computationally stable upper bound in the literature, combined with early stopping to prevent overfitting. Learning rate and sampling fractions were set to recommended values that produced consistent validation accuracy.

All models were evaluated using MAE, RMSE, and $R^2$ as defined in Section~\ref{sec:methodology}, with the final 7-day test set representing deployment.

\section{Results}\label{sec:results}

Model performance was evaluated using MAE, RMSE, and $R^2$, supported by visual inspection of forecasts.

\subsection{Facebook Prophet (FBP)}

FB Prophet effectively reproduced long-term seasonality and trend patterns, though short-term fluctuations were less precise. Test forecasts for PM$_{2.5}$ and PM$_{10}$ are shown in Fig.~\ref{fig:fbp_test}. Training forecast visualizations are provided in the Supplementary Material.

\begin{figure}[htbp]
    \centering
    \includegraphics[width=\linewidth]{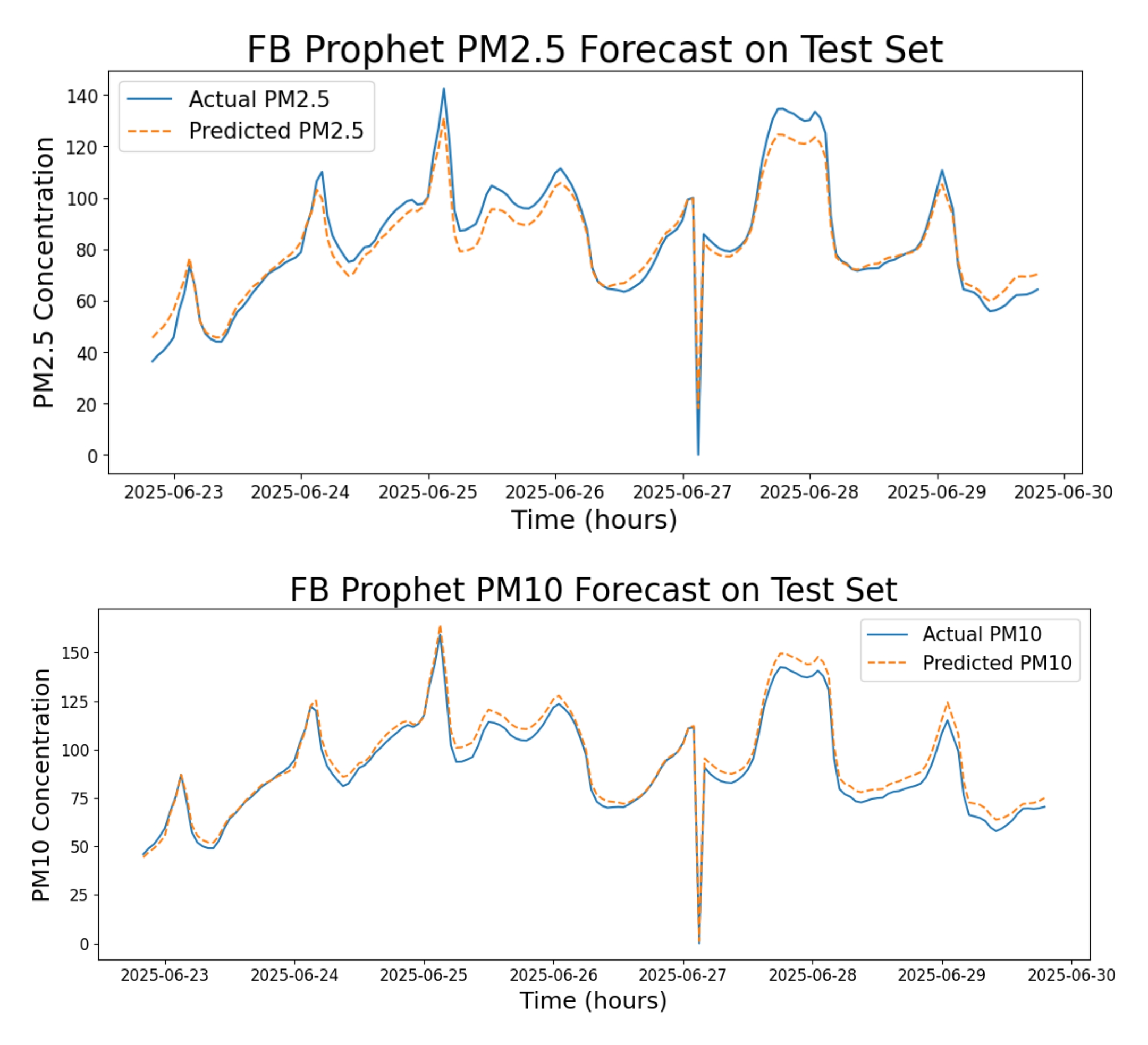}
    \caption{FB Prophet test forecasts (7-day horizon) for PM$_{2.5}$ and PM$_{10}$.}
    \label{fig:fbp_test}
\end{figure}

\subsection{NeuralProphet (NP)}

NeuralProphet captured seasonal and short-term autocorrelation through lagged regressors but produced higher errors than FBP, especially for PM$_{2.5}$. Testing forecast comparisons are shown in Fig.~\ref{fig:np_test}. Training forecast visualizations for each model are provided in the Supplementary Material. 

\begin{figure}[htbp]
    \centering
    \includegraphics[width=\linewidth]{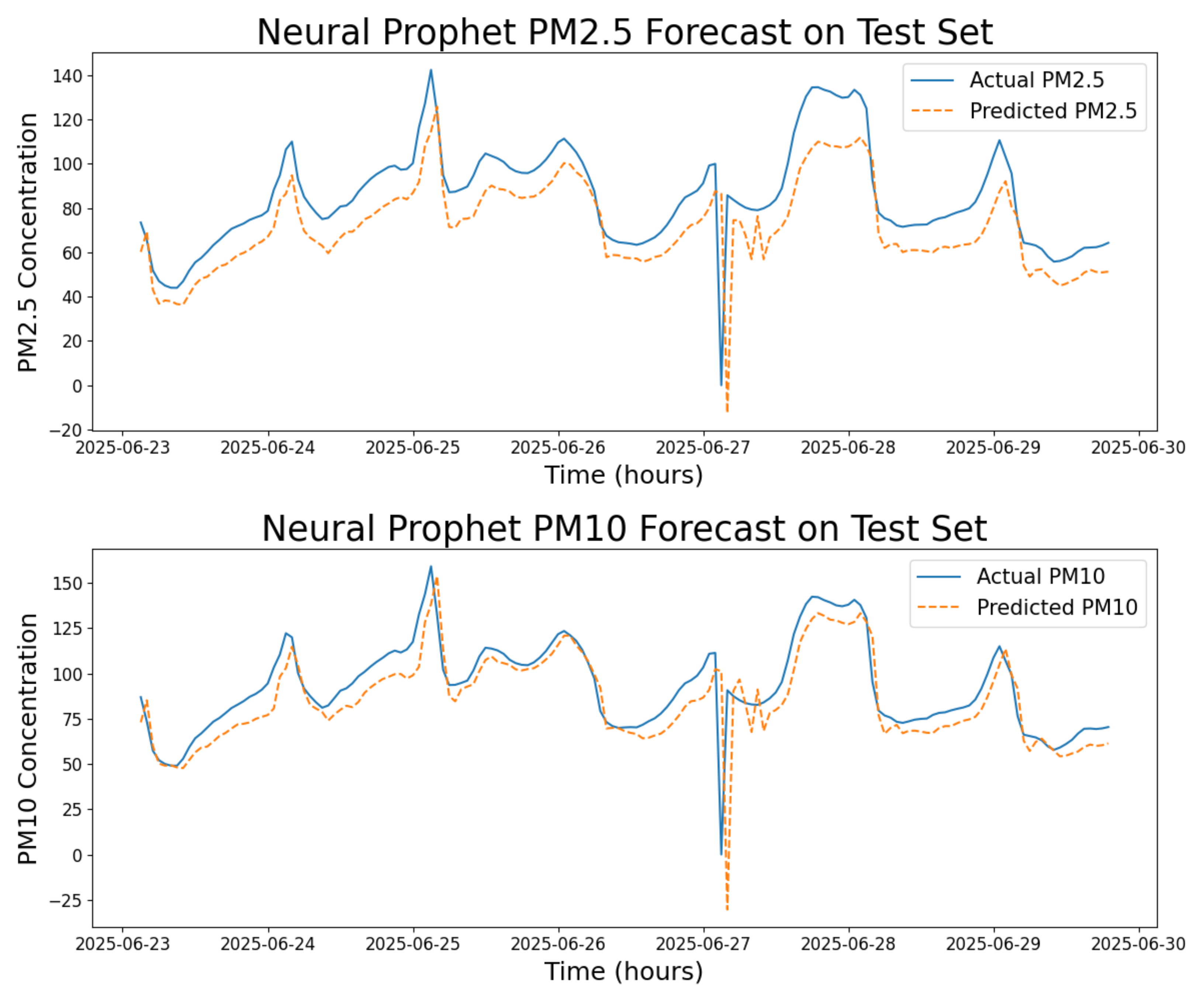}
    \caption{NeuralProphet test forecasts (7-day horizon) for PM$_{2.5}$ and PM$_{10}$.}
    \label{fig:np_test}
\end{figure}

\subsection{Additional Baselines: LSTM, SARIMAX, and LightGBM}

Among the baselines, LSTM captured sequential dependencies but showed limited generalization—moderate accuracy for PM$_{10}$ ($R^2=0.44$) and weak results for PM$_{2.5}$. LightGBM fit the training data strongly but overfitted, yielding negative $R^2$ on the test set. SARIMAX achieved reasonable accuracy as a traditional baseline but lagged behind Prophet-based models, reflecting its limited ability to model nonlinear seasonal effects.

\subsection{Quantitative Results}

The performance of all five models is summarized in Table~\ref{tab:all_results}. FB Prophet achieved the best generalization, with test $R^2$ above 0.94 for both pollutants. SARIMAX performed well as a statistical reference but lacked adaptability to complex seasonal patterns. NeuralProphet improved short-term dynamics but had higher variance, while LSTM and LightGBM showed weaker stability and overfitting tendencies.

\begin{table*}[htbp]
\centering
\caption{Quantitative comparison of training, validation, and test results using MAE, RMSE, and R² for both PM$_{2.5}$ and PM$_{10}$ targets.}
\label{tab:all_results}
\begin{tabular}{|l|l|c|c|c|c|c|c|c|}
\hline
Model & Target & \multicolumn{2}{c|}{Train} & \multicolumn{2}{c|}{Val} & \multicolumn{3}{c|}{Test} \\
 &  & MAE & RMSE & MAE & RMSE & MAE & RMSE & $R^2$ \\
\hline
FB Prophet & PM$_{2.5}$ & 14.4 & 24.6 & -- & -- & 4.3 & 5.4 & 0.94 \\
           & PM$_{10}$  & 14.5 & 30.3 & -- & -- & 4.0 & 4.6 & 0.96 \\
NeuralProphet & PM$_{2.5}$ & 29.2 & 52.0 & 24.3 & 41.9 & 14.6 & 17.8 & 0.38 \\
              & PM$_{10}$  & 33.3 & 60.4 & 29.0 & 50.3 & 10.0 & 16.1 & 0.56 \\
LSTM          & PM$_{2.5}$ & 26.8 & 49.7 & 28.5 & 49.6 & 38.8 & 40.0 & -2.49 \\
              & PM$_{10}$  & 26.4 & 55.3 & 26.6 & 57.3 & 10.6 & 17.6 & 0.44 \\
SARIMAX       & PM$_{2.5}$ & 2.7 & 6.9 & 15.8 & 31.1 & 17.1 & 18.0 & 0.41 \\
              & PM$_{10}$  & 3.0 & 8.7 & 19.3 & 39.7 & 13.7 & 14.9 & 0.64 \\
LightGBM      & PM$_{2.5}$ & 14.4 & 21.4 & 21.2 & 31.7 & 33.4 & 38.0 & -1.63 \\
              & PM$_{10}$  & 23.0 & 35.9 & 52.7 & 77.6 & 144.0 & 154.5 & -37.30 \\
\hline
\end{tabular}
\end{table*}

\subsection{Comparative Analysis}

\begin{figure}[htbp]
    \centering
    \includegraphics[width=\linewidth]{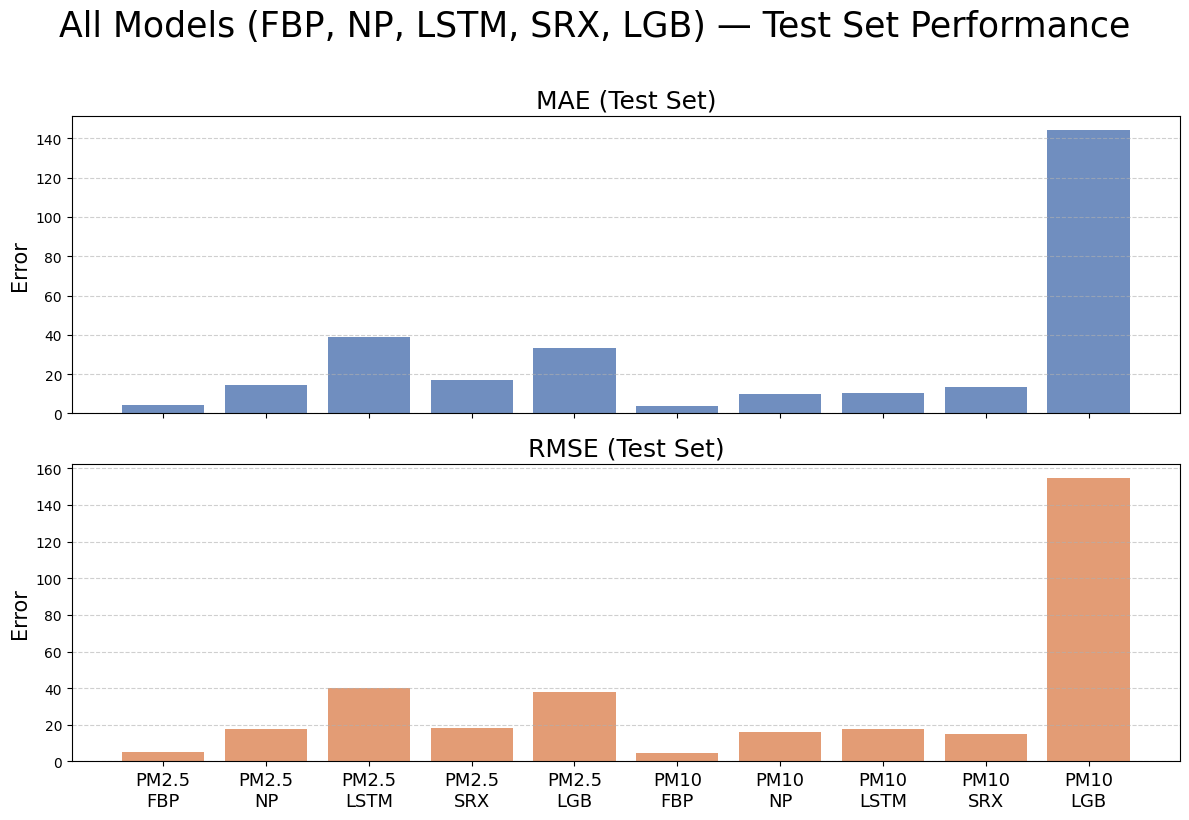}
    \caption{Bar charts of MAE and RMSE values summarize and contrast forecasting accuracy across all tested models.}
    \label{fig:compare}
\end{figure}

Figure~\ref{fig:compare} summarizes model performance via MAE and RMSE. FB Prophet consistently outperformed all others on the test horizon, maintaining the best balance of accuracy and interpretability. SARIMAX provided a stable but less flexible traditional baseline. NeuralProphet was moderately effective, while LSTM and LightGBM suffered from instability and overfitting.

Although the models were not explicitly benchmarked across pollution regimes, visual inspection of the multi-year series indicates stable performance trends across both high-pollution winter peaks and cleaner summer periods. Meteorological variables showed comparatively weak influence, with gaseous precursors remaining the dominant predictors. These observations align with the broader air-quality literature, where seasonal accumulation rather than short-term weather fluctuations typically drives PM variability.

A similar pattern was observed across hourly cycles: forecasting errors were higher during the afternoon peak pollution period, when pollutant concentrations rapidly increased, and more stable during off-peak hours. These trends align with boundary-layer dynamics, where limited mixing at night and early morning increases variability in pollutant levels. Despite these fluctuations, model performance rankings remained consistent. Figure S3 (Supplementary Material) illustrates the hourly variation in pollutant concentrations, showing the highest levels in the afternoon and the lowest in the early morning, likely due to traffic and industrial activities.

\section{Discussion}\label{sec:discussion}

All five models captured the seasonal and long-term dynamics of PM$_{2.5}$ and PM$_{10}$, with SARIMAX providing a stable but less adaptive statistical benchmark. FB Prophet achieved the strongest generalization, producing the lowest errors and highest $R^2$ on the test horizon, while NeuralProphet benefited from lagged terms but showed more variance. LSTM and LightGBM tended to overfit despite capturing short-term dependencies. Overall, lightweight additive models remained highly competitive with more complex approaches and offer a practical balance of accuracy, interpretability, and deployment efficiency.

Several limitations should be acknowledged. The models were not evaluated under large data gaps; while Prophet-based approaches are generally robust, substantial discontinuities could affect their stability. Informal testing indicated that removing key gaseous regressors led to a slight drop in accuracy, suggesting sensitivity to emission-related precursors. Furthermore, the results reflect Beijing’s emission profile and seasonal pollution cycles, and may not directly transfer to cities with different source compositions or meteorological regimes.

\section{Conclusion \& Future Work}\label{sec:conclusion}

This study evaluated lightweight additive models—Facebook Prophet and NeuralProphet—for forecasting PM$_{2.5}$ and PM$_{10}$ in Beijing, benchmarked against LSTM, SARIMAX, and LightGBM. Both additive frameworks successfully captured seasonal and long-term dynamics, with FB Prophet showing the most stable generalization on the 7-day horizon. The results demonstrate that accurate and interpretable forecasts can be achieved without the computational complexity of deep or large ensemble models, supporting their suitability for operational air-quality monitoring.

Looking ahead, our findings point to several practical directions for strengthening data-driven air-quality forecasting. A promising step is to combine data from heterogeneous sources—city monitors, roadside sensors, satellite estimates, and reanalysis products—within unified numerical pipelines, allowing models like Prophet to benefit from richer spatial and temporal signals. The study also highlights the need for improved data openness, as transparent APIs and publicly accessible pollutant records are crucial for reproducibility and for enabling wider community participation in air-quality analytics. Beyond data considerations, collaborative monitoring frameworks involving government agencies, research institutions, and environmental organizations could help extend model applicability from Beijing to other Chinese cities with distinct emission profiles. Finally, future work will explore hybrid techniques and federated learning schemes, enabling privacy-preserving model sharing across distributed monitoring networks while improving generalization across diverse urban environments.

\section*{Declarations}

\begin{itemize}
    \item \textbf{Funding:} No funding was received for conducting this research. 
    
    \item \textbf{Conflict of interest:} The authors declare no competing interests.
        
    \item \textbf{Availability of data and materials:} The dataset used in this study is available from publicly available APIs, as detailed in the Section~\ref{sec:implementation}.  
    The full code implementation is available at:  
    \href{https://github.com/moazzamumer/Air-Quality-Forecasting-Comparison-Research}{GitHub Repository}
    
\end{itemize}


\begingroup
\setlength\bibsep{0pt plus 0.2ex}     
\renewcommand*{\bibfont}{\footnotesize}       
\bibliography{sn-bibliography}
\endgroup


\end{document}